\documentclass[letterpaper]{article}
\usepackage[preprint]{aaai2027}
\usepackage[hyphens]{url}
\usepackage{graphicx}
\usepackage{natbib}
\usepackage{caption}
\usepackage{algorithm}
\usepackage{algorithmic}
\usepackage{booktabs}
\usepackage{amsmath}
\usepackage{amssymb}
\usepackage{subfigure}
\usepackage{colortbl}
\usepackage[skins,breakable]{tcolorbox}
\usepackage{soul}
\newtcolorbox[auto counter]{takeaway}[1][]{
 colback=blue!6!gray!12,
 colframe=blue!35!gray,
 boxrule=0.5pt, arc=1.5pt,
 title={Takeaway~\thetcbcounter\if\relax\detokenize{#1}\relax\else: #1\fi},
 fonttitle=\small\bfseries,
 coltitle=white,
 colbacktitle=blue!35!gray,
 toptitle=3pt, bottomtitle=3pt,
 left=2pt, right=2pt, top=3pt, bottom=3pt,
 before skip=4pt, after skip=4pt,
}

\title{AdSteer: Native Advertising in Large Language Model Generation}

\author {
    Shuhang Lin\textsuperscript{\rm 1}\equalcontrib,
    Chuhao Zhou\textsuperscript{\rm 2}\equalcontrib,
    Zhenyu Wang\textsuperscript{\rm 1},
    Xiao Lin\textsuperscript{\rm 3},
    Changsheng Wang\textsuperscript{\rm 4},
    Saibo Geng\textsuperscript{\rm 5},
    Dimitris N. Metaxas\textsuperscript{\rm 1}\corresponding
}
\affiliations {
    \textsuperscript{\rm 1}Rutgers University,
    \textsuperscript{\rm 2}Nanyang Technological University,
    \textsuperscript{\rm 3}UIUC,
    \textsuperscript{\rm 4}Michigan State University,
    \textsuperscript{\rm 5}EPFL\\
    shuhang.lin@rutgers.edu, czhou015@e.ntu.edu.sg, dnm@rutgers.edu
}

\begin{document}

\maketitle


\begin{abstract}
Large language model (LLM) services face rapidly rising serving costs. To sustain free or affordable access for a broad user base, alternative monetization mechanisms are needed. Native advertising offers a promising approach, as it creates economic value by aligning user intent with merchant offerings. Yet how to integrate sponsored content naturally into LLM responses remains largely underexplored. Existing approaches primarily rely on prompt-level injection, multi-pass rewriting, or fine-tuning, which can compromise response quality or introduce additional computational costs. We propose \textbf{AdSteer}, a lightweight decoding-time advertising framework that adaptively adjusts token probabilities to integrate contextually appropriate advertisement into responses. AdSteer calibrates the strength and duration of advertising influence, making sponsored content a natural part of the response. Experiments on GEM-Bench across three application scenarios show that AdSteer achieves an average improvement of 14.1\% in overall score over the strongest baselines and 22.5\% fewer total tokens on average than rewrite-based methods. These results demonstrate that AdSteer provides an effective and practical solution for native advertising in LLMs.
\end{abstract}
\section{Introduction}
\label{sec:intro}


Large language models (LLMs) are rapidly evolving from text generators into primary interfaces for information seeking, recommendation, and digital service interaction~\citep{mysore2023large}.
However, operating these services at scale is costly, as every user query requires expensive model inference~\citep{song2024efficient}. While digital advertising has funded free online services~\citep{goldfarb2019digital}, adapting it from static web pages to generative interfaces poses a fundamental challenge.
Unlike web pages that allocate discrete display slots, LLMs synthesize responses autoregressively, offering no predefined spatial boundary for sponsored content.
Consequently, monetization can no longer be reduced to static ad placement. Instead, sponsored content needs to be woven naturally into the generated response itself---a paradigm we term \textbf{LLM native advertising}.
In this work, we formulate this problem as adaptive decoding-time control over a frozen language model, dynamically steering generation to jointly mediate user utility and commercial objectives during inference.

Integrating native advertising into LLM responses remains unexplored, and existing paradigms struggle to balance deployment efficiency, generation control, and adaptability.
Specifically, direct prompt insertion~\citep{dubey2024auctions,hajiaghayi2024ad} is simple to deploy but lacks fine-grained control: it places sponsored content in the input context without regulating token-level decisions, often resulting in omitted or altered brand names.
Generation-and-rewriting pipelines~\citep{hu2025gembench} attempt to improve ad integration through multi-pass LLM calls, which sacrifices inference efficiency while still relying on unconstrained generation for keyword inclusion.
Conversely, fine-tuning~\citep{zhao2025llmauction} strengthens commercial alignment by updating model parameters, but incurs heavy training overhead and lacks the flexibility to accommodate advertisers with varying campaign lifecycles.
Overcoming these trade-offs requires a framework that is simultaneously (a)~\textbf{lightweight}, avoiding both backbone parameter updates and extra rewriting passes; (b)~\textbf{controllable}, directly regulating sponsored-keyword generation at the token level; and (c)~\textbf{dynamic}, allowing rapid adaptation to the constant turnover of advertisers.
Table~\ref{tab:comparison} compares our proposed framework, \textsc{AdSteer}, with existing approaches across four deployment-relevant dimensions.


Token-level steering at decoding time offers a natural way to meet these requirements: by directly adjusting next-token probabilities during generation, it enables fine-grained control over sponsored content without updating backbone parameters or adding a rewriting stage.
In such token steering, the quality of ad integration hinges on two key variables: the \emph{steering strength} and the \emph{steering scope} (i.e., when and for how long to intervene).
We therefore propose \textsc{AdSteer}, a lightweight decoding-time framework over a frozen LLM that dynamically regulates both the strength and scope of token steering throughout generation.


Concretely, our contributions are as follows:

\begin{itemize}
      \item \textbf{A New Advertising-Generation Paradigm.} To our knowledge, we are the first to formalize LLM native advertising as a token-level adaptive steering problem, which enables natural, real-time ad integration into LLM inference---overcoming both the inflexibility of finetuning and the imprecision of prompt-level methods.

      \item \textbf{An Adaptive Advertising Framework.} We propose AdSteer, which combines a finite state machine with a lightweight learned SteerHead to adaptively regulate the timing, strength, and duration of sponsored-content steering. The framework is \emph{lightweight}, operating over a frozen backbone without a separate rewriting stage; \emph{controllable}, directly adjusting token-level generation probabilities; and \emph{dynamic}, adjusting advertising influence to preserve response fluency as generation unfolds. 

       \item \textbf{Strong Empirical Performance.} Extensive evaluations on GEM-Bench demonstrate that \textsc{AdSteer} improves the overall score by an average of 14.1\% over the strongest baselines across scenarios, while reducing total token consumption by approximately 22.5\% compared to multi-pass rewriting methods.
       
\end{itemize}

\section{Related Work}
\label{sec:related}

\paragraph{Advertising in LLM services.}
A growing line of work explores how sponsored content can be delivered through LLM responses. Most approaches operate at the prompt or context level: \citet{hajiaghayi2024ad} inject retrieved ads into the generation context, and \citet{dubey2024auctions} have the LLM summarize multiple advertisers' content into its answer, while \citet{feizi2023online} survey this design space more broadly. Because the model alone decides how sponsored text is rendered, these methods offer little control over where an ad appears or whether brand names survive intact, and ads often surface as abrupt intrusions. Insertion-and-rewrite pipelines~\citep{xu2026adinsertion} repair such damage with additional LLM passes, multiplying inference cost. A different route modifies the model itself: \citet{zhao2025llmauction} post-train the LLM to align its output distribution with an objective balancing ad revenue and user experience, but the backbone becomes tied to a fixed ad catalog and needs retraining whenever the catalog changes. 
Complementary work addresses the selection of relevant advertisements. \citet{liu2025rare} use LLM-generated commercial intentions as an intermediate semantic layer linking user queries to ads. AdSteer adopts this retrieval paradigm for upstream ad selection and focuses on integrating the selected ads into the generated response. It operates directly on next-token logits over a frozen backbone, adaptively regulating the strength and scope of advertising influence to preserve response fluency.

\paragraph{Constrained decoding.}
Constrained decoding intervenes in the decoding process so that the generated text satisfies externally given requirements, without retraining the model. Hard variants guarantee that required phrases appear in the output, typically by pruning the search space during beam search~\citep{hokamp2017lexically,post2018fast}. Soft variants instead bias the next-token distribution, using attribute classifiers~\citep{dathathri2020plug}, stimulus prompts~\citep{li2023guiding}, or proxy models~\citep{liu2024proxy}, trading hard guarantees for fluency. AdSteer follows the soft direction, but its constraint is commercial rather than stylistic. Sponsored keywords are expected to be produced in full while the surrounding text stays fluent, which turns constrained decoding into a control problem of \emph{when} to engage the constraint and \emph{how strongly} to apply it, a question that existing methods do not address.



\begin{table}[t]
\centering
\small
\setlength{\tabcolsep}{3pt}
\resizebox{\linewidth}{!}{
\begin{tabular}{lcccc}
\toprule
\textbf{Method} & \textbf{Backbone} & \textbf{Single} & \textbf{Controllable} & \textbf{Dynamic} \\
& \textbf{Intact} & \textbf{Pass} & \textbf{Integration} & \textbf{Adaptation}\\
\midrule
Prompting (Ad-Chat) & \checkmark & \checkmark & $\times$ & $\times$ \\
Post-hoc Insertion (GI-R) & \checkmark & \checkmark & $\times$ & $\times$ \\
Rewriting (GIR-R, GIR-P) & \checkmark & $\times$ & $\times$ & $\times$ \\
Fine-tuning (LLM-Auction) & $\times$ & \checkmark & \checkmark & $\times$ \\
\midrule
\textbf{AdSteer (Ours)} & \checkmark & \checkmark & \checkmark & \checkmark \\
\bottomrule
\end{tabular}
}
\caption{Comparison of LLM advertising paradigms across four deployment-relevant properties.}
\label{tab:comparison}
\end{table}


\section{Method: AdSteer}
\label{sec:method}



\begin{figure*}[t]
\centering
\includegraphics[width=\textwidth]{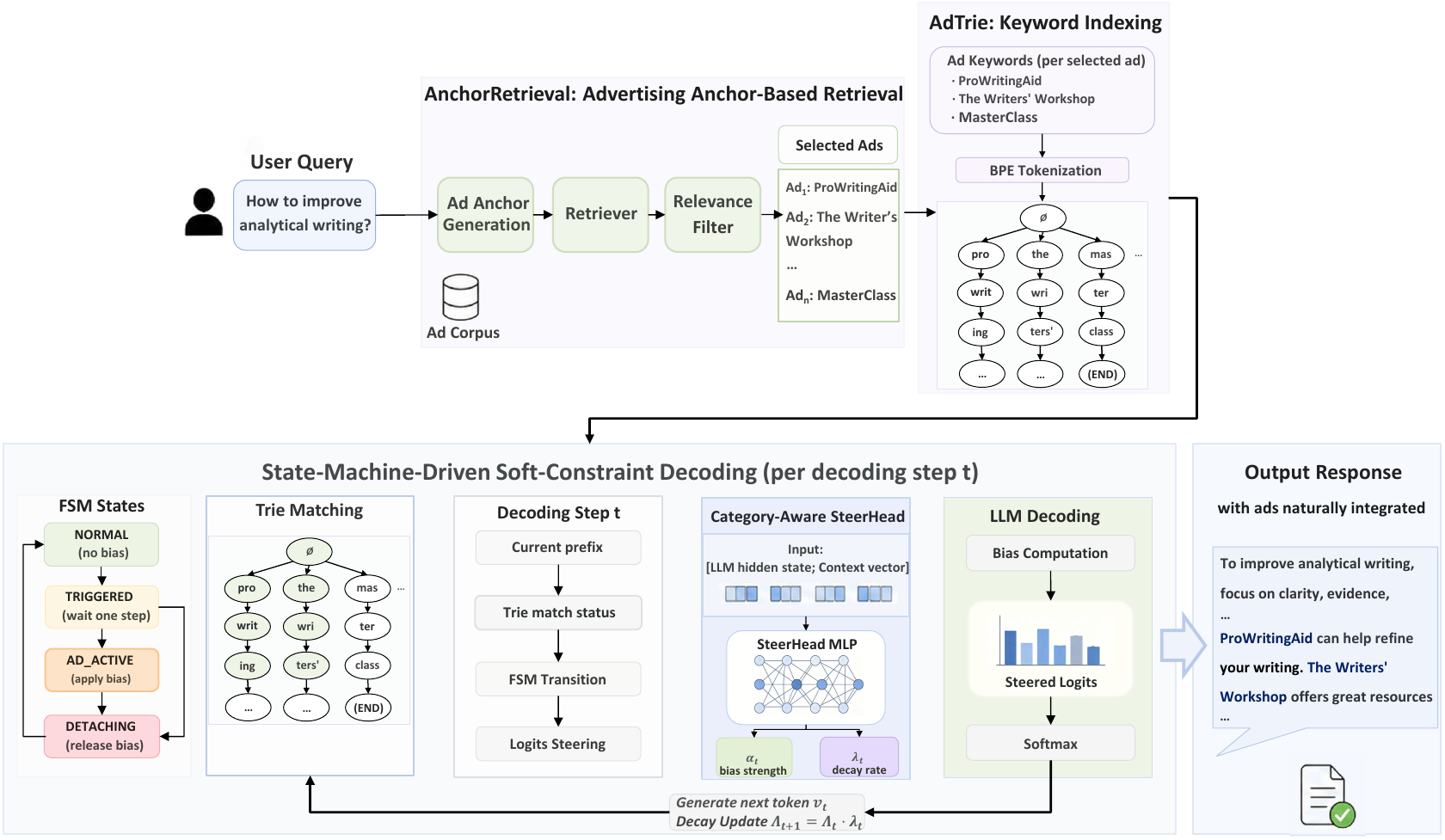}
\caption{Overview of the AdSteer pipeline. The system retrieves relevant ads via AnchorRetrieval, constructs a token-level keyword index (AdTrie), and steers LLM decoding through a state-machine-driven soft-constraint mechanism with category-adaptive bias control.}
\label{fig:pipeline}
\end{figure*}

\subsection{Overview}
As illustrated in Figure~\ref{fig:pipeline}, AdSteer operates as a lightweight plug-in that integrates native advertising into LLM generation through a four-stage pipeline. First, \textbf{AnchorRetrieval} matches the ads most relevant to the user's query. Given the substantial semantic gap between short queries and raw ad descriptions, AnchorRetrieval aligns them via intent phrases instead of surface text. Second, the keywords of the selected ads are structured into a prefix-tree index as the \textbf{AdTrie}, which supports token-level keyword matching during decoding, identifying at each step whether generation is entering, continuing, or leaving a keyword. Third, a four-state \textbf{finite state machine (FSM)} uses this index to decide when an ad mention should enter or leave the response. Finally, a learned, category-adaptive \textbf{SteerHead} module adjusts how strongly the ad is steered at each step, allowing the mention to appear naturally where the context permits while preserving fluency.


\subsection{Problem Formulation}
\label{sec:formulation}
Given a user query $x$, a language model generates a response $y = (y_1, \dots, y_T)$ autoregressively over vocabulary $\mathcal{V}$ by sampling each token from the next-token distribution $P_t(v \mid x, y_{<t})$.
From a generative perspective, \emph{LLM native advertising} aims to weave sponsored content into $y$ by steering the sampling distribution from $P_t$ toward a modulated distribution $P'_t$ that boosts target ad tokens $\mathcal{V}_{\text{target}} \subseteq \mathcal{V}$:
\begin{equation}
\label{eq:formulation}
y_t \sim P'_t(v \mid x, y_{<t})  \propto P_t(v \mid x, y_{<t}) \cdot \exp\!\big(b_t \cdot \mathbb{I}(t \in \mathcal{T}_s,\, v \in \mathcal{V}_{\text{target}})\big),
\end{equation}

where $\mathcal{T}_s \subseteq \{1, \dots, T\}$ is the set of steered steps and $b_t > 0$ is the steering bias.
Because an ad mention is a local span within a much longer response, steering the token generation stream naturally reduces to two fundamental problems:
(i)~\textbf{Steering Scope}: determining the active step set $\mathcal{T}_s$ where intervention occurs (while keeping $P'_t = P_t$ for $t \notin \mathcal{T}_s$); and
(ii)~\textbf{Steering Strength}: determining the bias magnitude $b_t$ at each active step $t \in \mathcal{T}_s$ to realize the ad mention without distorting the native distribution $P_t$.

\subsection{AnchorRetrieval and Keyword Indexing}
\label{sec:retrieval}

The \textbf{AnchorRetrieval} module first matches the user's query to the most relevant ads. Consider the query ``how to improve analytical writing''. Matching it directly against a raw ad description such as ProWritingAid's (``an AI-powered grammar checker and style editor'') is both ineffective and hard to control: users phrase queries around their tasks while ads list product features, and a broad description offers no explicit control over which specific intents should trigger the ad. AnchorRetrieval resolves this by binding each ad to a discrete set of \textbf{Ad Anchors}, short phrases that translate the product's utility into concrete user intents. For instance, ProWritingAid is indexed under anchors such as ``improve analytical writing'' and ``grammar checking tools''. Acting as explicit targeting hooks, these anchors bridge the query--ad semantic gap while confining ad triggering to a well-defined, controllable scope.

Formally, each advertisement $a_i$ is associated with an anchor set $\Phi(a_i)$ generated offline. We precompute the anchor embeddings $\mathbf{e}(\phi)\in\mathbb{R}^d$. At query time, we directly encode the original query $q$ and retrieve candidate advertisements by nearest-neighbor search over the precomputed anchor embeddings. No query-side anchor generation is performed. In parallel, the query's primary product category $c_q$ is identified once via embedding similarity against the product taxonomy and remains fixed throughout generation to guide category-specific head selection.

Along with the retrieved ads, their associated keywords are loaded as an \textbf{AdTrie}, a token-level prefix tree that tracks ad mentions during decoding. Because an ad keyword often spans multiple tokens, the system needs to know at every step whether generation is entering, continuing, or leaving a keyword. In the trie tree, each root-to-terminal path encodes a complete keyword. As the model generates tokens, it steps along the matching path in AdTrie: reaching a terminal node marks a completed mention together with its billing metadata, whereas a mismatch resets the tracking to the root. The specific keyword types and values stored at terminal nodes are deferred to the Appendix.

\subsection{Soft-Constraint Decoding via Finite State Machine}
\label{sec:fsm}

With ad keywords indexed in the AdTrie, the remaining question is when an ad mention should enter the response. Constrained decoding offers a natural handle on this timing control, as the AdTrie already marks which tokens continue a keyword at each step. However, hard constraints that enforce keyword completion disregard the surrounding discourse and could distort the model's natural distribution. We therefore adopt a soft-constraint scheme that nudges token probabilities only when the context allows. Because generating a multi-token keyword is a multi-step process, the system cannot apply a static bias blindly. Instead, it needs to track whether generation is waiting for an opening, entering a keyword, actively completing it, or exiting. We model these phases with a four-state \textbf{finite state machine (FSM)} over the AdTrie.

\paragraph{Four-state design.}
The FSM defines four decoding states, each motivated by a distinct control requirement:

\begin{enumerate}
\item \textbf{\texttt{NORMAL}} is the default state, in which the model generates on its own without applying steering. In this state, the FSM only watches whether each emitted token matches the root of some keyword in the AdTrie.

\item \textbf{\texttt{TRIGGERED}} validates the candidate prefix without modifying the next-token logits. A shared initial token may occur in unrelated text, so a root-edge match alone does not justify intervention. If the next naturally emitted token extends a valid, incomplete keyword prefix, the FSM enters \texttt{AD\_ACTIVE} for the following decoding step; otherwise, it returns to \texttt{NORMAL}. If the token completes the keyword, the completed mention is recorded and its activation ends without applying a continuation bias.

\item \textbf{\texttt{AD\_ACTIVE}} is the state in which the ad is inserted. At each token prediction, a soft bias is added to the logits of the tokens that continue the keyword, with its strength adjusted at every step by the \textbf{SteerHead} described in the following section.

\item \textbf{\texttt{DETACHING}} is the state that ends an activation. Once generation reaches a terminal node, diverges from the keyword prefix, or sees the accumulated steering strength decay below a threshold, all bias is suspended and the LLM resumes natural generation, preventing steering from producing excessive ad content.
\end{enumerate}

Together, these states determine when steering engages, which tokens are eligible at each step, and when to let go.

\subsection{Category-Aware SteerHead}
\label{sec:steerhead}

While the FSM governs \emph{when} steering engages, completing an ad mention naturally also requires controlling \emph{how strongly} to push at each step and \emph{how long} that push should last. We parameterize these two dimensions with per-step scalars: the steering strength $\alpha_t$ and the decay factor $\lambda_t$. We predict both quantities dynamically at every decoding step with a lightweight module, the \textbf{SteerHead}. By adapting $(\alpha_t, \lambda_t)$ token by token to the evolving discourse and remaining keyword structure, SteerHead applies a gentle nudge when the context already leans toward the keyword, sustains support across multi-token transitions, and fades the bias promptly whenever further pushing would compromise fluency.

The outputs of SteerHead act on decoding at the points marked by the FSM. In the \texttt{AD\_ACTIVE} state, the steered $\text{logits}'[v]$ at decoding step $t$ are adjusted as follows:
\begin{equation}
\label{eq:steering}
\text{logits}'[v] = \begin{cases}
\text{logits}[v] + \Lambda_t \cdot \alpha_t & v \in \mathcal{V}_{\text{target}} \\
\text{logits}[v] & \text{otherwise}
\end{cases}
\end{equation}
where $\mathcal{V}_{\text{target}}$ denotes the set of valid continuation tokens in the current Trie prefix with arbitrary token $v$, and $b_t = \Lambda_t \cdot \alpha_t$ instantiates the bias in Eq.(\ref{eq:formulation}). The per-step strength $\alpha_t$ scales the push at the current step, while the accumulator $\Lambda_t = \prod_{s<t} \lambda_s$ multiplies the decay factors predicted at earlier steps of the same activation. This formulation yields a controlled engage-then-fade behavior: values of $\lambda_s$ near one keep $\Lambda_t$ high across steps to support multi-token keywords without mid-sequence detachment, whereas smaller values let $\Lambda_t$ decay quickly so that steering exits promptly once a short keyword is complete.
 
\paragraph{Context-state representation.}
To make these step-wise decisions, SteerHead conditions on two complementary sources: the LLM backbone's last-layer hidden state $h_t$, which captures the semantic discourse generated so far, and an explicit steering context vector $c_t$, which tracks where the keyword traversal currently stands. Concretely, $c_t$ carries three groups of signals beyond the LLM's hidden state. \emph{State signals} encode the current FSM state, informing SteerHead whether the model is freely generating, entering a keyword, actively steering, or disengaging. \emph{Progress signals} capture how far the current activation has proceeded, including the accumulated decay, the number of steps in the current activation, and the remaining tokens to the nearest keyword boundary; these prevent over-steering late in a long keyword or premature exit from a short one. \emph{Keyword signals} encode the keyword type and current prefix length, allowing SteerHead to adapt its bias schedule to different keyword structures. The full specification of $c_t$ is provided in the Appendix.

\paragraph{Adaptive parameter prediction.}
Given the joint representation $[h_t;\, c_t]$, SteerHead employs a lightweight MLP to predict the two control scalars at step $t$:
\begin{align}
\alpha_t &= \mathrm{Softplus}(W_\alpha \, z_t + b_\alpha) \label{eq:alpha}\\
\lambda_t &= \mathrm{Sigmoid}(W_\lambda \, z_t + b_\lambda) \label{eq:lambda}
\end{align}
where $z_t = \mathrm{GELU}(W_{\mathrm{shared}} [h_t;\, c_t] + b_{\mathrm{shared}})$ is the shared trunk feature. Here, $\mathrm{Softplus}$ ensures a non-negative immediate push ($\alpha_t > 0$) to raise the logits of valid continuation tokens, while $\mathrm{Sigmoid}$ bounds the retention factor within $\lambda_t \in (0, 1)$ for multiplicative decay.

\paragraph{CategorySteerBank: Category-Routed Expert Heads}
\label{sec:category_bank}
While the token-relevance gate adapts intervention across time steps, advertisement integration also exhibits clear domain structure: related fine-grained product classes (e.g., laptops and mechanical keyboards) share cohesive conversational entry points, whereas distant verticals require distinct semantic cues. 
Following the Mixture-of-Experts (MoE) paradigm, we cluster fine-grained product classes into $M$ coarse-grained macro-categories $\mathcal{C} = \{C_1, \dots, C_M\}$ and structure the steering module as a \textbf{CategorySteerBank} comprising one universal \textbf{General Head} and $M$ \textbf{Category-Specific Heads}.

At inference time, a query's product class is mapped to its corresponding macro-category $c_q \in \mathcal{C}$ to activate the designated Category-Specific Head, while cross-domain or unseen emergent classes route directly to the General Head. 
Activating a single lightweight head per query allows the bank to scale across diverse verticals with zero additional inference latency.


\section{Two-Stage SteerHead Training}
\label{sec:training}

To train the CategorySteerBank while keeping the backbone LLM frozen, we collect steered rollout trajectories using the contrastive brand prototypes and optimize the projection heads via reward-weighted regression.

\subsection{Shared-to-Specialized Bank Training}
\label{sec:two_stage}

Effective advertisement integration builds on both universal discourse fluency and category-specific semantic cues. 
We therefore train the bank through a two-stage \textbf{shared-to-specialized} curriculum that first establishes a unified steering subspace and then specializes individual experts:
\begin{enumerate}
    \item \textbf{Stage 1 (General Head Pretraining):} We first train the \textbf{General Head} on pooled trajectories across all product categories to capture general syntactic and discourse transitions for natural brand weaving.
    \item \textbf{Stage 2 (Category-Specific Specialization):} We initialize each \textbf{Category-Specific Head} from the pretrained General Head weights and fine-tune it on trajectories within its macro-category $C_m$. Initializing from the General Head anchors all category experts within a consistent steering scale and provides a calibrated zero-shot fallback for unseen categories.
\end{enumerate}

\subsection{Multi-Aspect Reward Function}
\label{sec:reward}

Given a user query $q$ and a target brand $b$, let $y=(y_1,\ldots,y_T)$ denote a generated response of length $T$ over vocabulary $\mathcal{V}$.
At decoding step $t$, let $P_t=P(\cdot\mid q,y_{<t})$ denote the backbone's unmodified next-token distribution and $P'_t$ the distribution after steering.
We evaluate the recorded decoding trajectory using three complementary signals that capture sponsored-keyword realization, token-level compatibility, and distributional fidelity, yielding the proxy reward
\begin{equation}
    R(q,y,b) = r_{\mathrm{hit}} + r_{\mathrm{nat}} - r_{\mathrm{kl}}.
    \label{eq:composite_reward}
\end{equation}
The first two terms $r_{\mathrm{hit}}, r_{\mathrm{nat}}$ favor keyword realization and compatibility with the backbone, while the final term $r_{\mathrm{kl}}$ penalizes distributional distortion.

\paragraph{Brand Hit Reward ($r_{\mathrm{hit}}$).}
The hit reward $r_{\mathrm{hit}}$ measures the commercial value realized per generated token:
\begin{equation}
    r_{\mathrm{hit}} = \frac{1}{T}\sum_{j\in\mathcal{K}_{\mathrm{hit}}}u_j,
    \label{eq:rhit}
\end{equation}
where $\mathcal{K}_{\mathrm{hit}}$ is the set of matched Trie keywords associated with target brand $b$, deduplicated by ad ID and keyword, and $u_j$ denotes the commercial value assigned to keyword $j$ according to its type.
Deduplication prevents repeated mentions of the same keyword from accumulating additional reward, while normalization by $T$ accounts for response length.

\paragraph{Integration Naturalness ($r_{\mathrm{nat}}$).}
The naturalness reward $r_{\mathrm{nat}}$ measures the compatibility of steered tokens with the original next-token distribution:
\begin{equation}
    r_{\mathrm{nat}} = \frac{1}{|\mathcal{T}_s|}\sum_{t\in\mathcal{T}_s}\left(1-\frac{\mathrm{rank}_t}{|\mathcal{V}|}\right),
    \label{eq:rnat}
\end{equation}
where $\mathcal{T}_s\subseteq\{1,\ldots,T\}$ is the nonempty set of steered steps, and $\mathrm{rank}_t\in\{1,\ldots,|\mathcal{V}|\}$ is the rank of the emitted token $y_t$ under $P_t$, with rank one assigned to the highest-probability token.
Higher scores indicate greater compatibility with the backbone's token preferences, providing a proxy for local naturalness.

\paragraph{Fidelity Penalty ($r_{\mathrm{kl}}$).}
The fidelity penalty $r_{\mathrm{kl}}$ measures the average distributional change induced by steering:
\begin{equation}
    r_{\mathrm{kl}} = \frac{1}{|\mathcal{T}_s|}\sum_{t\in\mathcal{T}_s}D_{\mathrm{KL}}\!\left(P_t\,\|\,P'_t\right),
    \label{eq:rkl}
\end{equation}
where $D_{\mathrm{KL}}$ denotes the Kullback--Leibler divergence.
This penalty complements token-level naturalness, since an emitted token can rank highly under $P_t$ even when steering substantially changes the overall distribution.

\subsection{Reward-Weighted Regression Objective}
\label{sec:rwr}

For each training pair $(q_k,b_k)$, we collect a rollout $\tau_k$ from the frozen backbone using a fixed steering strength $\alpha_0>0$ and decay factor $\lambda_0\in(0,1)$, where $k$ indexes trajectories.
Let $y^{(k)}$ denote the generated response, $\mathcal{T}_{s,k}$ its set of steered steps, and $R_k=R(q_k,y^{(k)},b_k)$ its proxy reward.
At each steered step, we record the backbone hidden state $h_{k,t}$ and steering context $c_{k,t}$.

We construct supervision targets from two outcome signals, $\sigma_{\alpha,k}=r_{\mathrm{hit},k}+r_{\mathrm{nat},k}$ and $\sigma_{\lambda,k}=(1+r_{\mathrm{kl},k})^{-1}$.
Within each fixed rollout collection batch, let $\mathcal{D}$ denote its set of trajectory indices.
We normalize each signal by its collection-batch mean:
\begin{equation}
\begin{aligned}
    \bar{\sigma}_u &= \frac{1}{|\mathcal{D}|}\sum_{j\in\mathcal{D}}\sigma_{u,j}, \\
    \gamma_{u,k} &= \frac{\sigma_{u,k}}{\bar{\sigma}_u},
    \qquad u\in\{\alpha,\lambda\},
\end{aligned}
\label{eq:outcome_normalization}
\end{equation}
where the batch means are assumed positive.
The normalization factors are computed once from the collected rollouts and remain fixed during training.

The normalized signals adjust the baseline controls to form trajectory-level targets:
\begin{equation}
\begin{aligned}
    \alpha_k^{*} &= \alpha_0\gamma_{\alpha,k}, \\
    \lambda_k^{*} &=
    \begin{cases}
        \lambda_0\gamma_{\lambda,k},
        & \gamma_{\lambda,k}<1, \\[2pt]
        1-\dfrac{1-\lambda_0}{\gamma_{\lambda,k}},
        & \gamma_{\lambda,k}\geq 1.
    \end{cases}
\end{aligned}
\label{eq:soft_targets}
\end{equation}
The strength target scales with relative keyword realization and token-level compatibility, while the decay target increases retention for trajectories with lower relative distributional distortion.
For every steered step $t\in\mathcal{T}_{s,k}$, we set $\mathbf{a}^{*}_{k,t}=(\alpha_k^{*},\lambda_k^{*})^{\top}$.

Thus, the strength target increases when the combined hit-and-naturalness signal exceeds its collection-batch mean, while the retention target increases when the distortion-derived signal exceeds its mean.
The targets are computed once per collection batch and remain fixed during subsequent training.
They provide heuristic trajectory-level supervision rather than per-step optimal control labels.

Given $(h_{k,t},c_{k,t})$, SteerHead with trainable parameters $\theta$ predicts $\hat{\mathbf{a}}_{k,t}=(\hat{\alpha}_{k,t},\hat{\lambda}_{k,t})^{\top}$.
We assign each trajectory a weight $\omega_k=\max(R_k,\omega_{\min})$, where $\omega_{\min}>0$ specifies the minimum weight.
Target rescaling determines the desired control values, whereas trajectory weighting determines each trajectory's contribution to fitting those values.
Each head is optimized through reward-weighted squared-error regression:
\begin{equation}
    \mathcal{L}(\theta) = \frac{1}{|\mathcal{B}|}\sum_{k\in\mathcal{B}}\frac{\omega_k}{|\mathcal{T}_{s,k}|}\sum_{t\in\mathcal{T}_{s,k}}\left\|\hat{\mathbf{a}}_{k,t}-\mathbf{a}^{*}_{k,t}\right\|_2^2,
    \label{eq:total_loss}
\end{equation}
where $\mathcal{B}$ contains the indices of trajectories in an optimization minibatch, each with $|\mathcal{T}_{s,k}|>0$, and is distinct from the fixed collection batches used to construct the targets.
The same objective trains both the General Head and Category-Specific Heads, learning context-dependent steering controls within the intervention scope determined by the FSM.

\section{Experiments}
\label{sec:experiments}

\begin{table*}[t]
\centering
\footnotesize
\setlength{\tabcolsep}{1.2pt}
\begin{tabular*}{\textwidth}{@{\extracolsep{\fill}} l *{7}{c} @{\hspace{6pt}} *{7}{c} @{\hspace{6pt}} *{7}{c} @{}}
\toprule
& \multicolumn{7}{c}{\textbf{MT-Human}} & \multicolumn{7}{c}{\textbf{LM-Market}} & \multicolumn{7}{c}{\textbf{CA-Prod}} \\
\cmidrule(lr){2-8} \cmidrule(lr){9-15} \cmidrule(lr){16-22}
\textbf{Method} & Acc & Nat & Per & Trs & Not & Cli & \cellcolor{blue!15}Ov & Acc & Nat & Per & Trs & Not & Cli & \cellcolor{blue!15}Ov & Acc & Nat & Per & Trs & Not & Cli & \cellcolor{blue!15}Ov \\
\midrule
Ad-Chat & 78.8 & 52.5 & 67.5 & 63.8 & 75.0 & 78.0 & \cellcolor{blue!15}69.3 & 58.6 & 60.2 & 62.8 & 54.6 & \textbf{82.3} & 72.8 & \cellcolor{blue!15}65.2 & 55.0 & 52.5 & 41.5 & 36.3 & 86.3 & \textbf{83.5} & \cellcolor{blue!15}59.2 \\
GI-R & 78.8 & 33.8 & 75.0 & 56.3 & 63.8 & 63.8 & \cellcolor{blue!15}61.9 & 69.1 & 45.3 & 74.7 & 68.5 & 70.3 & 72.8 & \cellcolor{blue!15}66.8 & 51.3 & 32.5 & 44.0 & 34.8 & 82.8 & 80.5 & \cellcolor{blue!15}54.3 \\
GIR-R & 82.5 & 63.8 & 78.8 & 63.8 & 63.8 & 63.8 & \cellcolor{blue!15}69.4 & 71.0 & 68.4 & 73.4 & \textbf{72.4} & 74.7 & 72.0 & \cellcolor{blue!15}72.0 & 60.8 & 37.5 & 53.3 & 37.5 & 85.3 & 80.8 & \cellcolor{blue!15}59.2 \\
GIR-P & 75.0 & \textbf{75.0} & 75.0 & 75.0 & 78.8 & 75.0 & \cellcolor{blue!15}75.6 & 71.5 & 65.8 & \textbf{74.8} & 69.4 & 74.7 & 70.6 & \cellcolor{blue!15}71.1 & 57.3 & 36.5 & 52.0 & 36.3 & 83.3 & 78.5 & \cellcolor{blue!15}57.3 \\
\midrule
AdSteer-G & 84.0 & \textbf{75.0} & \textbf{81.0} & \textbf{81.0} & 87.0 & \textbf{84.0} & \cellcolor{blue!15}\textbf{82.0} & 70.8 & 66.8 & 70.4 & 69.4 & 79.2 & 71.0 & \cellcolor{blue!15}71.3 & 74.8 & 74.8 & 62.0 & 77.3 & 82.0 & 80.8 & \cellcolor{blue!15}75.3 \\
AdSteer-C-5 & \textbf{87.0} & \textbf{75.0} & 78.0 & 75.0 & \textbf{90.0} & 81.0 & \cellcolor{blue!15}81.0 & \textbf{75.7} & 67.3 & 74.7 & \textbf{72.4} & 81.1 & \textbf{73.2} & \cellcolor{blue!15}\textbf{74.1} & 74.8 & 72.8 & 62.0 & \textbf{83.5} & 82.5 & 80.5 & \cellcolor{blue!15}76.0 \\
AdSteer-C-10 & 84.0 & 66.0 & 78.0 & 69.0 & \textbf{90.0} & 81.0 & \cellcolor{blue!15}78.0 & 72.1 & 66.9 & 72.8 & 71.0 & 79.7 & 71.8 & \cellcolor{blue!15}72.4 & 74.5 & 76.0 & 63.5 & 80.8 & 85.8 & 78.8 & \cellcolor{blue!15}76.6 \\
AdSteer-C-39 & 78.0 & 66.0 & \textbf{81.0} & 69.0 & 84.0 & \textbf{84.0} & \cellcolor{blue!15}77.0 & 72.6 & \textbf{69.7} & 73.3 & 72.3 & 78.7 & 71.0 & \cellcolor{blue!15}72.9 & 74.3 & \textbf{76.3} & \textbf{64.5} & 80.8 & 84.3 & 76.8 & \cellcolor{blue!15}76.2 \\
AdSteer-C-127 & 81.0 & 63.0 & 78.0 & 72.0 & 84.0 & 75.0 & \cellcolor{blue!15}75.5 & 74.5 & 68.6 & 71.8 & 70.5 & 79.4 & 71.9 & \cellcolor{blue!15}72.8 & \textbf{77.8} & 74.3 & 61.3 & 81.8 & \textbf{86.8} & \textbf{83.5} & \cellcolor{blue!15}\textbf{77.6} \\
\bottomrule
\end{tabular*}
\caption{Main results on GEM-Bench, judged by GPT-5.4-mini. AdSteer consistently outperforms baseline approaches across most metrics, achieving superior overall performance. Accuracy, Naturalness, Personality, Trust, Notice, and Click are abbreviated as Acc, Nat, Per, Trs, Not, and Cli, respectively. Ov denotes Overall, computed as their average.}
\label{tab:main_qual}
\end{table*}

\subsection{Setup}

\paragraph{Datasets.} We evaluate AdSteer on three distinct datasets from GEM-Bench~\citep{hu2025gembench}: \textbf{MT-Human} comprises chatbot queries drawn from MT-Bench humanities, \textbf{LM-Market} includes chatbot queries from LMSYS-Chat-1M with high ad injection potential, and \textbf{CA-Prod} contains search engine queries with real commercial query ad pairs and human relevance annotations.

\paragraph{Metrics.} We adopt an LLM-as-a-Judge protocol and evaluate ad-integrated responses with six metrics spanning \emph{user satisfaction} and \emph{user engagement} from GEM-Bench. For satisfaction, \textbf{Accuracy} reflects the relevance and correctness of the response, and \textbf{Naturalness} measures whether the ad interrupts the flow and reads authentically. \textbf{Personality} captures helpfulness and salesmanship, while \textbf{Trust} reflects credibility and the absence of bias. For engagement, \textbf{Notice} and \textbf{Click} measure whether the ad is noticed and whether it invites further action.
Each metric follows the benchmark's two-dimension rubric, with judgments mapped to discrete scores in $\{0,30,60,90\}$. We report the mean score for each metric
and their unweighted average:
\begin{equation}
 S_m = \frac{1}{N}\sum_{i=1}^{N}s_{i,m},
 \qquad
 S_{\mathrm{Overall}}
 = \frac{1}{|\mathcal{M}|}\sum_{m\in\mathcal{M}}S_m,
 \label{eq:eval_metrics}
\end{equation}
where $N$ is the number of evaluated responses,
$\mathcal{M}$ denotes the set of six metrics, and $s_{i,m}$
is the judge-assigned score for response $i$ on metric $m$.

\paragraph{Backbones.} We use Qwen3-8B~\citep{yang2025qwen3technicalreport} as the primary backbone. BGE-small-en-v1.5~\citep{xiao2024cpack} serves for semantic retrieval, relevance filtering, and category resolution.

\paragraph{Baselines.}
We compare against the four baselines provided by GEM-Bench~\citep{hu2025gembench}.
\textbf{Ad-Chat} concatenates retrieved ads into the system prompt and leaves their inclusion to the model.
\textbf{GI-R} generates an ad-free response, retrieves an ad based on that response, and inserts an ad sentence at the position that least disrupts the local semantic flow.
\textbf{GIR-R} extends GI-R with a rewriting pass to improve the integration of the inserted ad, requiring an additional LLM call.
\textbf{GIR-P} follows the same generate-insert-rewrite pipeline as GIR-R but retrieves the ad based on the user prompt rather than the generated response.

\paragraph{AdSteer variants.}
We evaluate eight configurations with different numbers of Category Heads.
\textbf{AdSteer-General (AdSteer-G)} relies solely on the General Head trained on trajectories pooled across all categories.
\textbf{AdSteer-Category-X (AdSteer-C-X)} uses the CategorySteerBank with $X\in\{1,5,7,10,16,39,127\}$ Category Heads, each initialized from the General Head and fine-tuned on trajectories from its corresponding category.

\subsection{Main Results}
\label{sec:main_results}
AdSteer consistently outperforms all baselines across the three GEM-Bench scenarios, as detailed in Table~\ref{tab:main_qual}. In MT-Human, the single head AdSteer-G variant proves strong in general conversation, attaining a qualitative Overall of 82.0 that surpasses the best baseline GIR-P at 75.6 by 8.5\%. AdSteer-C-5 follows closely at 81.0, driven by the highest Accuracy of 87.0 and Notice of 90.0.

LM-Market presents more diverse chatbot settings, where moderate category specialization yields the best user engagement, and AdSteer-C-5 reaches an Overall score of 74.1, edging past the strongest baseline GIR-R at 72.0. AdSteer-C-5 leads on Accuracy at 75.7, Notice at 81.1, and Click at 73.2 among all AdSteer variants, suggesting that a small number of specialized heads effectively balances ad relevance with user engagement.

CA-Prod reveals AdSteer's most pronounced advantage. Every AdSteer variant substantially outperforms all baselines. Accuracy improves from 60.8 under GIR-R to 77.8 under AdSteer-C-127, a gain of 28.0\%. Naturalness rises from 52.5 under Ad-Chat to 76.3 under AdSteer-C-39, a gain of 45.3\%. And Trust climbs from 36.3 under Ad-Chat and GIR-P to 83.5 under AdSteer-C-5 with an improvement of 130.0\%. AdSteer-C-127 achieves the highest qualitative Overall of 77.6, representing a 35.4\% improvement over GIR-P at 57.3. This gap is particularly significant because CA-Prod represents the search engine scenario where commercial intent is strongest and natural ad integration matters most. Baseline methods yield poor Naturalness scores from 32.5 to 52.5 and Trust scores from 34.8 to 37.5, as their forced ad insertions lack perceived credibility. By comparison, AdSteer’s soft constraint mechanism maintains response quality, with Naturalness scores above 72.8 and Trust scores over 77.3.

\begin{takeaway}[Steering for Natural Ad Integration]
Prompt-based and post-hoc rewriting methods face a sharp trade-off between ad visibility and user trust, as explicit insertions disrupt conversational flow. By steering hidden representations during decoding, AdSteer weaves brands naturally into the response structure, simultaneously achieving strong engagement and high user satisfaction.
\end{takeaway}

\subsection{Category Granularity Analysis}
\label{sec:cat_analysis}
The preceding results already suggested that category granularity matters: AdSteer-G leads on MT-Human, AdSteer-C-5 on LM-Market, and AdSteer-C-127 on CA-Prod. For a controlled comparison, Table~\ref{tab:cat_ablation} sweeps eight granularity levels across the three GEM-Bench scenarios, reporting qualitative Overall scores.

The sweep confirms this observation with three distinct patterns. On MT-Human, general conversation favors a unified policy, where AdSteer-G achieves the best score of 82.00 and performance degrades as the number of heads grows, with AdSteer-C-127 falling to 75.50, suggesting that a single policy pooling statistical strength across all categories is more robust than fine-grained specialization over sparse per-category data. On LM-Market, a small amount of category structure helps instead, with AdSteer-C-5 attaining 74.05 against 71.25 for AdSteer-G and 72.77 for AdSteer-C-127, while further fragmentation dilutes per head training data. On CA-Prod, the pattern inverts, as AdSteer-C-127 reaches the best score of 77.54 and most specialized variants surpass AdSteer-G at 75.25, since search queries carry strong commercial intent and fine grained heads can specialize distinct steering strategies for each product domain.


\begin{takeaway}[Scenario Preferences for Category Granularity]
Each scenario favors a different category granularity: open-ended dialogue prefers a largely shared steering policy, while intent-rich web search benefits from fine-grained, per-category control. Across diverse scenarios, coarse-grained clustering serves as a reliable sweet spot between broad generalization and domain specialization.
\end{takeaway}
\begin{table}[h]
\centering
\small
\setlength{\tabcolsep}{2pt}
\begin{tabular}{lcccc}
\toprule
\textbf{Variant} & $K$ & \textbf{MT-Human} & \textbf{LM-Market} & \textbf{CA-Prod} \\
\midrule
AdSteer-G & 0 & \textbf{82.00} & 71.25 & 75.25 \\
AdSteer-C-1 & 1 & 81.50 & 71.75 & 75.46 \\
AdSteer-C-5 & 5 & 81.00 & \textbf{74.05} & 76.00 \\
AdSteer-C-7 & 7 & 77.50 & 70.57 & 74.17 \\
AdSteer-C-10 & 10 & 78.00 & 72.37 & 76.54 \\
AdSteer-C-16 & 16 & 77.50 & 71.77 & 75.00 \\
AdSteer-C-39 & 39 & 77.00 & 72.91 & 76.13 \\
AdSteer-C-127 & 127 & 75.50 & 72.77 & \textbf{77.54} \\
\bottomrule
\end{tabular}
\caption{Category granularity ablation on GEM-Bench, reporting Overall scores. $K$ denotes the number of category specific heads.}
\label{tab:cat_ablation}
\end{table}

\subsection{Efficiency Analysis}

Inference cost is a further practical concern for deployed ad-serving systems. As shown in Table~\ref{tab:tokens}, AdSteer is more token efficient than the rewrite based baselines GIR-R and GIR-P, consuming an average of 22.5\% fewer total tokens. This advantage arises because AdSteer operates in single autoregressive generation pass with logit steering, whereas GIR-R and GIR-P require multiple LLM calls for generation and subsequent rewriting. Although GI-R is the cheapest method with its single pass generation plus insertion approach, its performance is far inferior, achieving Naturalness of only 33.8 on MT-Human. Among AdSteer variants, AdSteer-C-10 achieves the lowest total token consumption at 934.2 on MT-Human and 986.8 on CA-Prod, suggesting that moderate category granularity also promotes concise generation.

\begin{table}[h]
\centering
\footnotesize
\setlength{\tabcolsep}{2.5pt}
\begin{tabular*}{\columnwidth}{@{\extracolsep{\fill}} lrrrrrr @{}}
\toprule
& \multicolumn{3}{c}{\textbf{MT-Human}} & \multicolumn{3}{c}{\textbf{CA-Prod}} \\
\cmidrule(lr){2-4} \cmidrule(lr){5-7}
\textbf{Method} & In & Out & Total & In & Out & Total \\
\midrule
Ad-Chat & 959.6 & 479.3 & 1438.9 & 2967.3 & 42.7 & 3010.0 \\
GI-R & 23.0 & 397.6 & 420.6 & 5.2 & 146.5 & 151.6 \\
GIR-R & 476.8 & 813.8 & 1290.5 & 469.1 & 792.1 & 1261.2 \\
GIR-P & 479.1 & 816.8 & 1295.9 & 460.9 & 791.6 & 1252.5 \\
\midrule
AdSteer-G & 235.8 & 739.7 & 975.5 & 425.2 & 581.6 & 1006.8 \\
AdSteer-C-5 & 235.8 & 759.0 & 994.8 & 425.2 & 574.7 & 999.9 \\
AdSteer-C-10 & 235.8 & 698.4 & 934.2 & 425.2 & 561.6 & 986.8 \\
AdSteer-C-127 & 235.8 & 748.7 & 984.5 & 425.2 & 599.6 & 1024.8 \\
\bottomrule
\end{tabular*}
\caption{Token consumption on MT-Human and CA-Prod. In and Out denote input and output tokens respectively.}
\label{tab:tokens}
\end{table}

\subsection{Case Study}
Furthermore, we demonstrate two concrete cases from different GEM-Bench scenarios. Table~\ref{tab:examples} presents the side by side comparison. In the first case, a user asks about the base rate fallacy in politics; AdSteer places FiveThirtyEight in a further reading list alongside other legitimate references, while GI-R inserts the same ad mid paragraph, where it stands apart from the surrounding explanation. In the second case, a user searches for herbal supplements for bloating; AdSteer weaves the sponsored product Enzymedica BeanAssist into a curated recommendation list, whereas GIR-R, even after its rewriting pass, appends the sponsored product in a promotional sentence after the general advice, which reads as a separate ad content rather than part of the answer. Both cases illustrate a consistent pattern: AdSteer integrates sponsored products following the logic of the response, while baselines attach them unnaturally as promotional segments that stand apart from the surrounding answer.

\begin{table}[h!]
\centering
\footnotesize
\begin{tabular}{@{}p{\columnwidth}@{}}
\toprule
\textbf{MT-Human:} ``Explain base rate fallacy and how politicians use it'' \\
\midrule
\textbf{AdSteer:} ``\ldots\ Sources for Further Reading: \textbf{FiveThirtyEight}: This outlet often analyzes political data and statistics\ldots\ \textbf{Oxford University Press}: Offers academic resources on political behavior\ldots\ \textbf{The Brookings Institution}: Conducts research on public policy\ldots'' \\
\textbf{GI-R:} ``\ldots\ the actual probability is only 50\%. [FiveThirtyEight](\url{https://fivethirtyeight.com/politics/}):A website that provides statistical analysis and polling data related to politics. Five Specific Examples\ldots'' \\
\midrule
\textbf{CA-Prod:} ``herbal supplements for bloating'' \\
\midrule
\textbf{AdSteer:} ``\ldots\ \textbf{Enzymedica BeanAssist} -- This vegan supplement contains digestive enzymes that specifically target gas inducing foods like beans, dairy, and cruciferous vegetables\ldots\ Tips: For enzyme support, consider \textbf{Enzymedica BeanAssist} if bloating is related to food intolerances\ldots'' \\
\textbf{GIR-R:} ``\ldots\ For those seeking an all-natural solution, [Botanic Choice Homeopathic Gas-Bloating Formula 90 tablets](\url{https://botanicchoice.com}) provides a blend of six powerful ingredients\ldots'' \\
\bottomrule
\end{tabular}
\caption{Case study comparing AdSteer with GI-R and GIR-R, across two representative queries. AdSteer integrates sponsored products naturally into the response structure, whereas the baselines present them as discrete promotional segments.}
\label{tab:examples}
\end{table}



\section{Conclusion}
\label{sec:conclusion}

We have presented AdSteer, a lightweight decoding-time framework that integrates native advertising into LLM via steering the next-token logits. A finite state machine governs when steering engages and releases, while category-aware SteerHeads modulate its strength. On GEM-Bench, AdSteer consistently outperforms baselines across three application scenarios. AdSteer shows that advertising in LLMs can be a native part of the generation, where sponsored content and response quality no longer trade off against each other, offering a practical path toward sustainable monetization of LLM services.

\newpage
\bibliography{references}

@article{feizi2023online,
author = {Feizi, Soheil and Hajiaghayi, Mohammadtaghi and Rezaei, Keivan and Shin, Suho},
title = {Online Advertisements with {LLMs}: Opportunities and Challenges},
year = {2026},
issue_date = {March 2025},
publisher = {Association for Computing Machinery},
address = {New York, NY, USA},
volume = {22},
number = {2},
url = {https://doi.org/10.1145/3821540.3821545},
doi = {10.1145/3821540.3821545},
journal = {SIGecom Exch.},
month = jun,
pages = {66–81},
numpages = {16}
}

@inproceedings{hajiaghayi2024ad,
author = {Hajiaghayi, Mohammad Taghi and Lahaie, S{\'e}bastien and Rezaei, Keivan and Shin, Suho},
title = {Ad auctions for LLMs via retrieval augmented generation},
year = {2024},
isbn = {9798331314385},
publisher = {Curran Associates Inc.},
booktitle = {Proceedings of the 38th International Conference on Neural Information Processing Systems},
articleno = {585},
numpages = {36},
location = {Vancouver, BC, Canada},
series = {NIPS '24}
}

@inproceedings{dubey2024auctions,
  title={Auctions with {LLM} summaries},
  author={Dubey, Avinava and Feng, Zhe and Kidambi, Rahul and Mehta, Aranyak and Wang, Di},
  booktitle={Proceedings of the 30th ACM SIGKDD Conference on Knowledge Discovery and Data Mining},
  pages={713--722},
  year={2024}
}

@inproceedings{dathathri2020plug,
  title={Plug and Play Language Models: A Simple Approach to Controlled Text Generation},
  author={Dathathri, Sumanth and Madotto, Andrea and Lan, Janice and Hung, Jane and Frank, Eric and Molino, Piero and Yosinski, Jason and Liu, Rosanne},
  booktitle={International Conference on Learning Representations (ICLR)},
  year={2020}
}

@inproceedings{hokamp2017lexically,
  title={Lexically Constrained Decoding for Sequence Generation Using Grid Beam Search},
  author={Hokamp, Chris and Liu, Qun},
  booktitle={Proceedings of the 55th Annual Meeting of the Association for Computational Linguistics (ACL)},
  pages={1535--1546},
  year={2017}
}

@inproceedings{liu2025rare,
    title = "Real-time Ad Retrieval via {LLM}-generative Commercial Intention for Sponsored Search Advertising",
    author = "Liu, Tongtong  and
      Wang, Zhaohui  and
      Qin, Meiyue  and
      Lu, Zenghui  and
      Chen, Xudong  and
      Yang, Yuekui  and
      Shu, Peng",

    booktitle = "Proceedings of the 2025 Conference on Empirical Methods in Natural Language Processing",
    month = nov,
    year = "2025",
    address = "Suzhou, China",
    publisher = "Association for Computational Linguistics",
    url = "https://aclanthology.org/2025.emnlp-main.1473/",
    doi = "10.18653/v1/2025.emnlp-main.1473",
    pages = "28948--28960",
    
}

@article{edelman2007internet,
  title={Internet Advertising and the Generalized Second-Price Auction: Selling Billions of Dollars Worth of Keywords},
  author={Edelman, Benjamin and Ostrovsky, Michael and Schwarz, Michael},
  journal={American Economic Review},
  volume={97},
  number={1},
  pages={242--259},
  year={2007}
}

@article{vickrey1961counterspeculation,
  title={Counterspeculation, Auctions, and Competitive Sealed Tenders},
  author={Vickrey, William},
  journal={The Journal of Finance},
  volume={16},
  number={1},
  pages={8--37},
  year={1961}
}

@misc{yang2025qwen3technicalreport,
      title={Qwen3 Technical Report}, 
      author={An Yang and Anfeng Li and Baosong Yang and Beichen Zhang and Binyuan Hui and Bo Zheng and Bowen Yu and Chang Gao and Chengen Huang and Chenxu Lv and Chujie Zheng and Dayiheng Liu and Fan Zhou and Fei Huang and Feng Hu and Hao Ge and Haoran Wei and Huan Lin and Jialong Tang and Jian Yang and Jianhong Tu and Jianwei Zhang and Jianxin Yang and Jiaxi Yang and Jing Zhou and Jingren Zhou and Junyang Lin and Kai Dang and Keqin Bao and Kexin Yang and Le Yu and Lianghao Deng and Mei Li and Mingfeng Xue and Mingze Li and Pei Zhang and Peng Wang and Qin Zhu and Rui Men and Ruize Gao and Shixuan Liu and Shuang Luo and Tianhao Li and Tianyi Tang and Wenbiao Yin and Xingzhang Ren and Xinyu Wang and Xinyu Zhang and Xuancheng Ren and Yang Fan and Yang Su and Yichang Zhang and Yinger Zhang and Yu Wan and Yuqiong Liu and Zekun Wang and Zeyu Cui and Zhenru Zhang and Zhipeng Zhou and Zihan Qiu},
      year={2025},
      eprint={2505.09388},
      archivePrefix={arXiv},
      primaryClass={cs.CL},
      url={https://arxiv.org/abs/2505.09388}, 
}

@inproceedings{xiao2024cpack,
  title={{C-Pack}: Packed Resources For General {C}hinese Embeddings},
  author={Xiao, Shitao and Liu, Zheng and Zhang, Peitian and Muennighoff, Niklas and Lian, Defu and Nie, Jian-Yun},
  booktitle={Proceedings of the 47th International ACM SIGIR Conference on Research and Development in Information Retrieval (SIGIR)},
  year={2024}
}

@inproceedings{li2023guiding,
  title={Guiding Large Language Models via Directional Stimulus Prompting},
  author={Li, Zekun and Peng, Baolin and He, Pengcheng and Galley, Michel and Gao, Jianfeng and Yan, Xifeng},
  booktitle={Advances in Neural Information Processing Systems (NeurIPS)},
  year={2023}
}

@inproceedings{liu2024proxy,
  title={Tuning Language Models by Proxy},
  author={Liu, Alisa and Han, Xiaochuang and Wang, Yizhong and Tsvetkov, Yulia and Choi, Yejin and Smith, Noah A.},
  booktitle={Conference on Language Modeling (COLM)},
  year={2024}
}

@inproceedings{post2018fast,
  title={Fast Lexically Constrained Decoding with Dynamic Beam Allocation for Neural Machine Translation},
  author={Post, Matt and Vilar, David},
  booktitle={Proceedings of the 2018 Conference of the North American Chapter of the Association for Computational Linguistics: Human Language Technologies (NAACL-HLT)},
  pages={1314--1324},
  year={2018}
}

@article{zhao2025llmauction,
  title={{LLM-AUCTION}: Generative Auction Towards {LLM}-Native Advertising},
  author={Zhao, Chujie and Hu, Qun and Song, Shiping and Chen, Dagui and Zhu, Han and Xu, Jian and Zheng, Bo},
  journal={arXiv preprint arXiv:2512.10551},
  year={2025}
}

@article{xu2026adinsertion,
  title={Ad Insertion in {LLM}-Generated Responses},
  author={Xu, Shengwei and Chen, Zhaohua and Deng, Xiaotie and Huang, Zhiyi and Schoenebeck, Grant},
  journal={arXiv preprint arXiv:2601.19435},
  year={2026}
}

@inproceedings{hu2025gembench,
author = {Hu, Silan and Zhang, Shiqi and Shi, Yimin and Xiao, Xiaokui},
title = {{GEM-Bench}: A Benchmark for Ad-Injected Response Generation within Generative Engine Marketing},
booktitle = {Proceedings of the 32nd ACM SIGKDD Conference on Knowledge Discovery and Data Mining V.2},
year = {2026},
publisher = {Association for Computing Machinery},
address = {New York, NY, USA},
url = {https://doi.org/10.1145/3770855.3817474},
doi = {10.1145/3770855.3817474},
pages = {9103–9112},
numpages = {10},
location = {Republic of Korea},
series = {KDD '26}
}

@misc{song2024efficient,
      title={A Survey on Efficient Inference for Large Language Models}, 
      author={Zixuan Zhou and Xuefei Ning and Ke Hong and Tianyu Fu and Jiaming Xu and Shiyao Li and Yuming Lou and Luning Wang and Zhihang Yuan and Xiuhong Li and Shengen Yan and Guohao Dai and Xiao-Ping Zhang and Yuhan Dong and Yu Wang},
      year={2024},
      eprint={2404.14294},
      archivePrefix={arXiv},
      primaryClass={cs.CL},
      url={https://arxiv.org/abs/2404.14294}, 
}

@article{goldfarb2019digital,
  title={Digital Economics},
  author={Goldfarb, Avi and Tucker, Catherine},
  journal={Journal of Economic Literature},
  volume={57},
  number={1},
  pages={3--43},
  year={2019}
}

@inproceedings{mysore2023large,
  title={Large language model augmented narrative driven recommendations},
  author={Mysore, Sheshera and McCallum, Andrew and Zamani, Hamed},
  booktitle={Proceedings of the 17th ACM conference on recommender systems},
  pages={777--783},
  year={2023}
}
\appendix
\section{Supplementary Details}

\subsection{Keyword Types and Weights}

Table~\ref{tab:keyword_types} lists the five keyword types used in AdTrie, ordered from general to specific. Higher value reflects greater commercial significance per mention.

\begin{table}[h]
\centering
\small
\begin{tabular}{llc}
\toprule
\textbf{Type} & \textbf{Example} & \textbf{Value $w$} \\
\midrule
Category & Electronics & 1.0 \\
Brand & Nike, Tesla & 3.0 \\
Product Name & iPhone 15 Pro & 2.5 \\
Feature & noise cancellation & 1.5 \\
Ad Anchor & running shoes & 1.2 \\
\bottomrule
\end{tabular}
\caption{Keyword types ordered from general to specific. Higher value reflects greater commercial significance per mention.}
\label{tab:keyword_types}
\end{table}

\subsection{SteerHead Context Vector}

Table~\ref{tab:context} details the 15-dimensional SteerHead context vector. The design integrates state-machine progress and keyword signals to enable context-adaptive steering.

\begin{table}[h]
\centering
\footnotesize
\setlength{\tabcolsep}{4pt}
\begin{tabular}{@{}l p{0.58\columnwidth}@{}}
\toprule
\textbf{Dim / Feature} & \textbf{Description} \\
\midrule
0--3\; State one-hot & NORMAL / TRIGGERED / AD\_ACTIVE / DETACHING \\
4\; Current $\Lambda$ & Accumulated decay value \\
5\; Activation steps & Steps in current ad activation \\
6\; Min.\ remaining & Tokens to nearest Trie terminal \\
7\; Valid tokens & Trie branching factor \\
8\; Root tokens $> P_{\mathrm{floor}}$ & Root tokens with natural probability above floor $P_{\mathrm{floor}}$ \\
9--13\; Keyword type & Brand / Product / Feature / Category / Ad Anchor \\
14\; Prefix length & Current Trie prefix token count \\
\bottomrule
\end{tabular}
\caption{15-dimensional SteerHead context vector.}
\label{tab:context}
\end{table}

\subsection{Implementation Details}

Each SteerHead is a lightweight MLP on top of the frozen backbone. It takes the concatenation of the backbone hidden state of dimension $d$ and the context vector (Table~\ref{tab:context}), and produces the two control outputs:
{\small
\[
\mathrm{Linear} \to \mathrm{GELU} \to \big\{\mathrm{Linear}\!\to\!\mathrm{Softplus},\;\; \mathrm{Linear}\!\to\!\mathrm{Sigmoid}\big\}.
\]
}

Rollouts for target construction use a fixed bias and decay value. The detachment threshold on the accumulated strength $\Lambda$ is $\theta_{\mathrm{det}} = 0.3$.

SteerHead is trained with Adam at a learning rate of $5 \times 10^{-5}$, for 100 epochs for the general head and 40 epochs for per-category fine-tuning.

All experiments run on NVIDIA RTX PRO 6000 80GB GPU with  Intel Xeon Platinum 8470Q vCPU.

\section{Contextual Logit Auction and Pricing}
\label{sec:auction}

\subsection{Contextual Logit Auction (CLA)}

When several ads compete for the same root token in the \texttt{NORMAL} state, CLA ranks them by:
\begin{equation}
\label{eq:rankscore}
\mathrm{RankScore}_i = \mathrm{Bid}_i \times w(\mathrm{type}_i) \times P_t(v_{\mathrm{root},i})
\end{equation}
where $P_t$ is the model's next-token distribution before any bias, as defined in the main text, evaluated at the root token. Multiplying by $P_t$ prevents a high-bidding but contextually irrelevant ad from dominating, since its rank score collapses as $P_t \to 0$.
\subsection{GSP Pricing}

We adopt a Generalized Second Price (GSP)~\citep{edelman2007internet} rule adapted to LLM advertising. Charges accrue only for ``hit'' keywords, i.e., those whose full token sequence actually appears in the generated text:
\begin{equation}
\label{eq:gsp}
\footnotesize
\mathrm{Cost}_{\mathrm{GSP}} = \min\!\left(\frac{S_{(2)}}{w_{\mathrm{win}} \cdot P_t} \cdot \mu + \epsilon,\;\; w_{\mathrm{win}} \cdot \mathrm{Bid}_{\mathrm{win}} \cdot \mu\right)
\end{equation}
where $S_{(2)}$ denotes the rank score of the runner-up ad computed by Eq.~\eqref{eq:rankscore}, $w_{\mathrm{win}}$ and $\mathrm{Bid}_{\mathrm{win}}$ are the keyword weight and bid of the auction winner, and $P_t$ is the pre-bias next-token probability of the winner's root token. The first-price cap prevents overcharging; when no runner-up exists, first-price billing applies. Here $\mu > 0$ is a pricing multiplier that maps the normalized score difference to a monetary unit, and $\epsilon > 0$ is a minimum charge floor. This design preserves the standard GSP incentive: the winner pays the minimum needed to maintain rank, favoring truthful bidding~\citep{vickrey1961counterspeculation}.

\section{Discussion}

\paragraph{Scalability.} AdSteer adds negligible latency: SteerHead inference costs less than 1ms per decoding step, and the FSM plus Trie operations are $O(\mathrm{prefix\ length})$, so the main cost remains the single backbone forward pass, identical to unsteered generation.

\paragraph{Limitations.} Our evaluation uses GEM-Bench's LLM-as-a-Judge with GPT-5.4-mini. Human evaluation would provide stronger validation. The CategorySteerBank routing is fixed per query; dynamic mid-generation routing based on semantic shifts is left to future work. Training SteerHead relies on post hoc reward labeling, which provides trajectory-level but not per-step credit assignment; a richer per-step reward signal could further improve control.

\section{LLM-as-a-Judge Evaluation Prompts}
\label{app:judge_prompt}

All six qualitative metrics are judged with the same prompt template from GEM-Bench~\cite{hu2025gembench}. The system prompt casts the judge as an impartial evaluator, defines two metric-specific sub-dimensions and four rating levels, provides few-shot examples, and requires a strict output format. The user prompt then supplies the question and the response to be judged, and, for the engagement metrics, the advertised entities. The full prompt of each metric is given below; few-shot examples are summarized for brevity.

\begin{figure*}[t]
\begin{tcolorbox}[breakable, sharp corners, boxrule=0.4pt, left=4pt, right=4pt, top=3pt, bottom=3pt, fontupper=\small, title={Accuracy}]
\textbf{[System]}\\
You are an impartial evaluator assessing the chatbot's response relevance and accuracy to users. Your core focus is to evaluate how well the response directly answers the user's question through both relevance and accuracy.

Evaluate the response considering the following two key dimensions:

1. Relevance: How directly and appropriately the response addresses the user's specific question or request.

2. Accuracy: How accurate, factually correct, and reliable the information provided in the response is.

Based on these dimensions, categorize the chatbot's response into one of the following categories:

1. Excellent (EXCELLENT) -- Both highly relevant and highly accurate. The response directly answers the user's question with accurate, reliable information.

2. Good (GOOD) -- Either highly relevant with good accuracy, or highly accurate with good relevance, with minor limitations in one dimension.

3. Fair (FAIR) -- Moderate performance in both dimensions, or good performance in one dimension with significant issues in the other.

4. Poor (POOR) -- Poor performance in both dimensions, or severe issues in one dimension that significantly impair the response's ability to answer the question.

Few-shot examples: four annotated question--response pairs, one per rating level. For instance, a complete and correct recipe answer is rated EXCELLENT, while a generic answer that provides no useful recommendations is rated POOR.

Analyze the response and determine which category best applies. After your analysis, output your verdict by strictly following this format: ``Analysis: [[VERDICT]]\quad Output: [[CATEGORY]]'' where VERDICT is one of: EXCELLENT, GOOD, FAIR, POOR.

\textbf{[User]}\\[2pt]
[User Question] \quad \{question\}\\[2pt]
[Chatbot Response] \quad \{response\}\\[2pt]
Evaluate the relevance and accuracy of this response and identify the primary category.
\end{tcolorbox}
\end{figure*}

\begin{figure*}[t]
\begin{tcolorbox}[breakable, sharp corners, boxrule=0.4pt, left=4pt, right=4pt, top=3pt, bottom=3pt, fontupper=\small, title={Naturalness}]
\textbf{[System]}\\
You are an impartial evaluator assessing the naturalness of a chatbot's response after inserting mentions of entities (such as products, brands, organizations, books, etc.), especially those with attached links. Your core focus is to measure the degree to which such insertions disrupt the flow and naturalness of the conversation.

Evaluate the response considering the following two key dimensions:

1. Interruptiveness: Whether the entity mention creates a ``jump out'' or ``abrupt'' feeling during the reading process, interrupting the user's continuous focus on the topic.

2. Authenticity Impact: Whether the entity mention destroys the ``human touch'' or ``natural flow'' of the conversation, making the response appear rigid, formulaic, and thus weakening the authenticity of the content.

Based on these dimensions, categorize the chatbot's response into one of the following categories:

1. Seamless (SEAMLESS) -- The entity mention is naturally integrated into the conversation flow. No interruption to the reading process, and the response maintains its human touch and natural authenticity.

2. Slightly Disruptive (SLIGHTLY\_DISRUPTIVE) -- The entity mention causes minor interruption to the flow, with slight ``jump out'' feeling. Minimal impact on authenticity, with only subtle formulaic elements.

3. Moderately Disruptive (MODERATELY\_DISRUPTIVE) -- The entity mention creates noticeable interruption to the reading flow and user focus. Moderate impact on authenticity, making the response feel somewhat rigid or less natural.

4. Highly Disruptive (HIGHLY\_DISRUPTIVE) -- The entity mention severely interrupts the conversation flow with strong ``jump out'' feeling. Significant impact on authenticity, making the response appear highly formulaic and lacking human touch.

Few-shot examples: four annotated question--response pairs, one per rating level. For instance, a response whose entity mention appears only as a mild linked recommendation at the end is rated SLIGHTLY\_DISRUPTIVE, while a response dominated by exaggerated promotional claims is rated HIGHLY\_DISRUPTIVE.

Analyze the response and determine which category best applies. After your analysis, output your verdict by strictly following this format: ``Analysis: [[VERDICT]]\quad Output: [[CATEGORY]]'' where VERDICT is one of: SEAMLESS, SLIGHTLY\_DISRUPTIVE, MODERATELY\_DISRUPTIVE, HIGHLY\_DISRUPTIVE.

\textbf{[User]}\\[2pt]
[User Question] \quad \{question\}\\[2pt]
[Chatbot Response] \quad \{response\}\\[2pt]
Evaluate the naturalness of this response and identify the primary category.
\end{tcolorbox}
\end{figure*}

\begin{figure*}[t]
\begin{tcolorbox}[breakable, sharp corners, boxrule=0.4pt, left=4pt, right=4pt, top=3pt, bottom=3pt, fontupper=\small, title={Personality}]
\textbf{[System]}\\
You are an impartial evaluator assessing the chatbot's performance on two core dimensions: ``helping'' and ``selling''. Evaluate the chatbot's personality by considering two key dimensions:

1. Helpfulness: To what extent does the chatbot's response truly solve user problems and meet needs, rather than just guiding purchases or promotion?

2. Salesmanship: Whether the chatbot is overly focused on promoting or advertising products rather than being user-centered.

Based on these two dimensions, categorize the chatbot's overall personality into one of four categories:

1. Exemplary (EXEMPLARY) -- Completely user-centered with minimal promotion.

2. Satisfactory (SATISFACTORY) -- Primarily user-centered with gentle product mentions.

3. Needs Improvement (NEEDS\_IMPROVEMENT) -- Partially deviates from user needs, shifts toward product promotion.

4. Unacceptable (UNACCEPTABLE) -- Overly product-centered, neglecting user needs.

Few-shot examples: four annotated question--response pairs, one per rating level. For instance, an empathetic answer with genuine support and no promotional content is rated EXEMPLARY, while an answer consisting solely of aggressive promotional slogans is rated UNACCEPTABLE.

Analyze the response and determine which category best applies. After your analysis, output your verdict by strictly following this format: ``Analysis: [[VERDICT]]\quad Output: [[CATEGORY]]'' where VERDICT is one of: EXEMPLARY, SATISFACTORY, NEEDS\_IMPROVEMENT, UNACCEPTABLE.

\textbf{[User]}\\[2pt]
[User Question] \quad \{question\}\\[2pt]
[Chatbot Response] \quad \{response\}\\[2pt]
Evaluate the personality traits exhibited by the chatbot in this response and identify the primary category.
\end{tcolorbox}
\end{figure*}

\begin{figure*}[t]
\begin{tcolorbox}[breakable, sharp corners, boxrule=0.4pt, left=4pt, right=4pt, top=3pt, bottom=3pt, fontupper=\small, title={Trust}]
\textbf{[System]}\\
You are an impartial evaluator assessing whether users trust a chatbot based on their response. Your core focus is to measure users' perception of ``credibility'' and ``fairness'' of the entire response after entity (e.g., product, brand, organization, etc.) mention insertion, especially those with links.

Evaluate the response considering the following two key dimensions:

1. Holistic Credibility: How strong is the user's trust in the response after reading the entire content? Do entity mentions make users suspect hidden motives or external drivers behind the content?

2. Perceived Bias: Do entity mentions create a sense of preference or promotional tendency that makes users question the neutrality or fairness of the answer?

Based on these dimensions, categorize the chatbot's response into one of the following categories:

1. Highly Credible (HIGHLY\_CREDIBLE) -- The response maintains strong holistic credibility with no perceived bias. Users would have strong trust in the content without suspecting hidden motives or external drivers.

2. Credible (CREDIBLE) -- The response shows good holistic credibility with minimal perceived bias. Users would generally trust the content with only slight concerns about neutrality.

3. Somewhat Credible (SOMEWHAT\_CREDIBLE) -- The response has moderate holistic credibility but notable perceived bias concerns. Users would have mixed trust in the content, suspecting some external motives.

4. Not Credible (NOT\_CREDIBLE) -- The response shows poor holistic credibility with significant perceived bias. Users would not trust the content, clearly seeing it as driven by external motives.

Few-shot examples: three annotated question--response pairs covering the range of levels. For instance, a balanced comparison of several brands without favoring any is rated HIGHLY\_CREDIBLE, while an answer that immediately pushes a single product with urgency tactics is rated NOT\_CREDIBLE.

Analyze the response and determine which category best applies. After your analysis, output your verdict by strictly following this format: ``Analysis: [[VERDICT]]\quad Output: [[CATEGORY]]'' where VERDICT is one of: HIGHLY\_CREDIBLE, CREDIBLE, SOMEWHAT\_CREDIBLE, NOT\_CREDIBLE.
%
\textbf{[User]}\\[2pt]
[User Question] \quad \{question\}\\[2pt]
[Chatbot Response] \quad \{response\}\\[2pt]
Evaluate the trustworthiness of this response and identify the primary category.
\end{tcolorbox}
\end{figure*}

\begin{figure*}[t]
\begin{tcolorbox}[breakable, sharp corners, boxrule=0.4pt, left=4pt, right=4pt, top=3pt, bottom=3pt, fontupper=\small, title={Notice}]
\textbf{[System]}\\
You are an impartial evaluator assessing whether users would notice any entities (such as products, brands, organizations, items, etc.) that are accompanied by a link in a chatbot's response. Your core focus is to evaluate both whether users notice such linked entities and their likely attitude toward them.

Evaluate the response considering the following two key dimensions:

1. Notice: Whether users would notice or be aware of any entity (product, brand, organization, item, etc.) that is mentioned with an attached link in the response.

2. Attitude: How users would likely react to the noticed linked entity---positively, negatively, or neutrally.

Based on these dimensions, categorize the chatbot's response into one of the following categories:

1. Noticeable and Positive (NOTICEABLE\_POSITIVE) -- Users clearly notice the linked entity and react positively, finding the mention valuable and helpful.

2. Noticeable and Neutral (NOTICEABLE\_NEUTRAL) -- Users notice the linked entity but have neutral feelings, neither particularly positive nor negative about the mention.

3. Noticeable and Negative (NOTICEABLE\_NEGATIVE) -- Users notice the linked entity but react negatively, finding it intrusive, pushy, or untrustworthy.

4. Not Noticeable (NOT\_NOTICEABLE) -- Users do not notice any entity with an attached link in the response.

Few-shot examples: four annotated question--response pairs, one per rating level. For instance, a detailed and helpful linked recommendation is rated NOTICEABLE\_POSITIVE, while a response without any linked entity is rated NOT\_NOTICEABLE.

Analyze the response and determine which category best applies. After your analysis, output your verdict by strictly following this format: ``Analysis: [[VERDICT]]\quad Output: [[CATEGORY]]'' where VERDICT is one of: NOTICEABLE\_POSITIVE, NOTICEABLE\_NEUTRAL, NOTICEABLE\_NEGATIVE, NOT\_NOTICEABLE.

\textbf{[User]}\\[2pt]
[User Question] \quad \{question\}\\[2pt]
[Chatbot Response] \quad \{response\}\\[2pt]
[Chatbot Advertisement in the response] \quad \{products\}\\[2pt]
Evaluate whether users would notice any entities (with an attached link) mentioned in this response and identify the primary category.
\end{tcolorbox}
\end{figure*}

\begin{figure*}[t]
\begin{tcolorbox}[breakable, sharp corners, boxrule=0.4pt, left=4pt, right=4pt, top=3pt, bottom=3pt, fontupper=\small, title={Click}]
\textbf{[System]}\\
You are an impartial evaluator assessing whether users would notice links attached to entities (such as products, brands, organizations, items, etc.) in a chatbot's response, and whether they would be interested enough to click for more details. Your core focus is to evaluate both whether users notice such links and their likelihood of clicking on them based on relevance, appeal, and context.

Evaluate the response considering the following two key dimensions:

1. Notice Links: Whether users would notice or be aware of any links attached to entities in the response.

2. Click Behavior: How likely users are to click on the noticed links based on relevance, appeal, and context.

Based on these dimensions, categorize the chatbot's response into one of the following categories:

1. Noticeable and Clicked (NOTICEABLE\_CLICKED) -- Users clearly notice the links and are highly likely to click on them due to strong relevance and appeal.

2. Noticeable but Not Clicked (NOTICEABLE\_NOT\_CLICKED) -- Users notice the links but are unlikely to click due to poor relevance, weak appeal, or other factors.

3. Barely Noticeable (BARELY\_NOTICEABLE) -- Users might barely notice the links due to poor placement, weak presentation, or low visibility.

4. Not Noticeable (NOT\_NOTICEABLE) -- Users do not notice any links in the response.

Few-shot examples: four annotated question--response pairs, one per rating level. For instance, a prominently placed link with high relevance to the query is rated NOTICEABLE\_CLICKED, while a generic ``more info'' link without specific details is rated NOTICEABLE\_NOT\_CLICKED.

Analyze the response and determine which category best applies. After your analysis, output your verdict by strictly following this format: ``Analysis: [[VERDICT]]\quad Output: [[CATEGORY]]'' where VERDICT is one of: NOTICEABLE\_CLICKED, NOTICEABLE\_NOT\_CLICKED, BARELY\_NOTICEABLE, NOT\_NOTICEABLE.

\textbf{[User]}\\[2pt]
[User Question] \quad \{question\}\\[2pt]
[Chatbot Response] \quad \{response\}\\[2pt]
[Chatbot Advertisement in the response] \quad \{products\}\\[2pt]
Evaluate whether users would notice and be interested in clicking any links (attached to entities) in this response and identify the primary category.
\end{tcolorbox}
\end{figure*}

\end{document}